\documentclass[letterpaper, 10pt, conference]{ieeeconf}
\IEEEoverridecommandlockouts

\usepackage{graphicx}
\usepackage{cite}
\usepackage{amsmath,amssymb}
\usepackage{hyperref}
\usepackage{xcolor} 
\usepackage{comment}
\usepackage{algorithm}
\usepackage{algorithmic}

\usepackage{multirow}  
\usepackage{tcolorbox} 
\usepackage{array}     
\usepackage{overpic}

\newcommand{\removed}[1]{{\color{gray} }}

\title{OmniVIC: A Self-Improving Variable Impedance Controller with Vision-Language In-Context Learning for Safe Robotic Manipulation}

\author{Heng Zhang$^{*1}$, Wei-Hsing Huang$^{*2}$, Gokhan Solak$^{1}$, and Arash Ajoudani$^{1}$   
\thanks{This work was supported by the Horizon Europe Project TORNADO (GA 101189557).
}
\thanks{$^{*}$These two authors contributed equally to this work.}%
\thanks{$^{1}$Human-Robot Interfaces and Interaction Lab, Istituto Italiano di Tecnologia, Genova, Italy.}%
\thanks{$^{2}$Georgia Institute of Technology, Atlanta, USA.}%
}

\begin{document}

\maketitle

\begin{abstract}
We present OmniVIC, a universal variable impedance controller (VIC) enhanced by a vision language model (VLM), 
which improves safety and adaptation in any contact-rich robotic manipulation task to enhance safe physical interaction. 
Traditional VIC have shown advantages when the robot physically interacts with the environment, but lack generalization in unseen, complex, and unstructured safe interactions in universal task scenarios involving contact or uncertainty. 
To this end, the proposed OmniVIC interprets task context-derived reasoning from images and natural language and generates adaptive impedance parameters for a VIC controller. 
Specifically, the core of OmniVIC is a self-improving Retrieval-Augmented Generation(RAG) and in-context learning (ICL), 
where RAG retrieves relevant prior experiences from a structured memory bank to inform the controller about similar past tasks, and ICL leverages these retrieved examples and the prompt of the current task to query the VLM for generating context-aware and adaptive impedance parameters for the current manipulation scenario.
Therefore, a self-improved RAG and ICL guarantee OmniVIC works in universal task scenarios. OmniVIC facilitates zero-shot adaptation to a broad spectrum of previously unseen scenarios, even when the inputs of VLM are not well-represented.
The impedance parameter regulation is further informed by real-time force/torque feedback to ensure interaction forces remain within safe thresholds. 
We demonstrate that our method outperforms baselines on a suite of complex contact-rich tasks, both in simulation and on real-world robotic tasks, with improved success rates and reduced force violations. 
OmniVIC moves toward unifying high‑level semantic reasoning with low‑level compliant control,  and 
can be extended to safer, more generalizable manipulation across embodiments from robot arms to humanoids. 
Code, video and RAG dataset are available at 
\url{https://sites.google.com/view/omni-vic}
\end{abstract}


\section{Introduction}
Robotic manipulation in contact-rich environments presents significant challenges, particularly in ensuring safe and effective interactions with objects and the environment~\cite{ tang2023industreal,tsuji2025survey}. Tasks such as wiping a surface, inserting a plug into a socket, or pushing a drawer require not only precise motion planning but also adaptive compliance to handle uncertainties and variations in contact dynamics. 
While traditional variable impedance control (VIC) can improve safety and adaptability in specific scenarios by modulating the robot's stiffness and damping to achieve compliance, 
they typically require manual tuning of parameters and do not generalize well across different tasks or environments~\cite{narang2022factory}. 
Even learning-based VIC approaches~\cite{zhang2024srl} often struggle with generalization due to the limited scope of training data and the complexity of contact dynamics.

The emergence of VLMs has opened new avenues for enhancing robotic manipulation by providing rich semantic understanding and reasoning capabilities~\cite{11128409,11106445}. 
By leveraging the contextual information provided by VLMs, robots can better understand the nuances of a task, such as the required level of force or compliance, and adjust their behavior accordingly.
However, most existing VLM-based approaches focus on high-level task planning or trajectory generation, often neglecting the critical aspect of force-aware compliant control in contact-rich scenarios. 
This gap limits the effectiveness in ensuring safe and adaptive physical interactions, as they do not directly inform the low-level control strategies needed for such tasks.

While VLMs alone struggle to inform compliant control, retrieval-augmented generation (RAG) can fill this gap by recalling relevant past experiences, which has shown promise in enhancing model capabilities by allowing access to external knowledge bases at inference time~\cite{lewis2020retrieval}. 
In robotics, RAG can enable robots to retrieve relevant prior experiences or demonstrations that inform their current actions, improving adaptability and generalization~\cite{xie2024embodiedrag}. 
By integrating RAG with VLMs, robots can leverage both semantic understanding and experiential knowledge to better handle the complexities of contact-rich manipulation tasks.

Moreover, ICL enables robots to adapt behaviors from contextual examples without retraining~\cite{zhang2025dynamics}. 
Originally developed for language models, ICL allows models to perform new tasks by conditioning on demonstrations provided in the input prompt. 
In robotics, this concept has been extended to enable rapid task adaptation through visual prompts~\cite{zhu2024incoro,liu2024moka}, language instructions~\cite{liu2024vision}, or recent interaction experiences~\cite{gao2024physically}.
However, existing ICL approaches in robotics often focus on visual or trajectory demonstrations alone, without considering the physical interaction forces that are crucial for safe manipulation.

To address these challenges, we propose OmniVIC, a universal variable impedance controller enhanced by a VLM and empowered by RAG and ICL. 
OmniVIC integrates high-level semantic reasoning from multimodal inputs with low-level compliant control, enabling robots to adapt their impedance parameters for safe and effective manipulation in contact-rich scenarios. 
The key contributions of this work are:
\begin{itemize}
    \item We introduce a new paradigm of VIC that enables context-aware impedance adaptation by context-aware VLM for safe physical interaction. 
    OmniVIC allows robots to leverage semantic understanding and commonsense reasoning capabilities to inform low-level control strategies.
    
    \item We develop a self-improving RAG mechanism that retrieves relevant prior experiences from a structured memory bank, 
    facilitating the impedance adaptation
    based on similar past tasks. 
    This approach can locally enhance generalization and adaptability across diverse manipulation scenarios.
    
    \item We implement an ICL strategy that conditions the VLM on retrieved examples and current task context, 
    allowing for dynamic generation of impedance parameters without retraining. 
    This enables rapid adaptation to new tasks and environments.
    
    \item A suite of contact-rich manipulation tasks was conducted, demonstrating the effectiveness of OmniVIC in both simulation and real-world experiments. 
    Our results show that the proposed method not only enhances task success rates but also reduces force violations, enabling safer manipulation.  

\end{itemize}

\section{Related Work}
\subsection{Variable impedance control for physical tasks}
VIC has been widely recognized as a fundamental approach for enabling safe and adaptive robot behavior in contact-rich tasks \cite{tsuji2025survey,wei2025hmc}. Early works demonstrated that learning variable stiffness improves robustness and safety compared to fixed-gain control \cite{buchli2011learning} and that humans naturally adapt impedance to maintain stability in both stable and unstable interactions \cite{yang2011human}, providing strong biological motivation. More recently, reinforcement learning has leveraged VIC as an action space, improving performance in complex manipulation tasks \cite{bogdanovic2020learning}. Furthermore, it is shown to improve safety in training and deployment in learning-based approaches \cite{zhang2024srl}.



\subsection{Retrieval-Augmented generation for robotics}

RAG augments parametric models with non‑parametric memory, enabling inference‑time access to external knowledge~\cite{lewis2020retrieval}. In robotics, this allows agents to query prior embodied experience at run time rather than relying solely on retraining~\cite{xie2024embodiedrag,zhu2024raea}.

Robotic RAG differs from text‑only settings: retrieval must account for evolving spatio‑temporal context~\cite{booker2024embodiedrag} and often span multiple embodied modalities (e.g., vision and robot state signals)~\cite{xie2024embodiedrag}. Behavior Retrieval, in turn, shows that querying datasets for task‑relevant behaviors improves performance~\cite{du2023behavior}~\cite{lin2025flowretrieval}.

To the best of our knowledge, no previous studies have applied RAG to VIC in the context of contact-rich manipulation. 
To address this research gap, we propose a novel RAG-based approach for variable impedance control that uses prior experience without retraining while maintaining experience‑informed compliance; as the database grows, the robot improves its regulation of impedance parameters for novel contact‑rich scenarios.

\subsection{In-context learning}
Recent works have explored various approaches to implement ICL in robotic systems. Zhang et al.~\cite{zhang2025dynamics} leverage ICL for dynamics adaptation, while InCoro~\cite{zhu2024incoro} uses in-context learning for robot control with visual prompts. Other notable works include manipulate-anything~\cite{duan2024manipulate} which conditions policies on demonstration videos, and MOKA~\cite{liu2024moka} which combines ICL with meta-learning for rapid adaptation. VLM-based approaches~\cite{sarch2024vlm, liu2024vision} have shown promise in grounding language instructions through visual context, while~\cite{gao2024physically} explores physically-grounded ICL for manipulation tasks.
Existing methods largely focus on high-level planning or trajectory generation, overlooking force-aware compliant control in contact-rich tasks. 

Our method uses force-aware ICL with retrieval-augmented examples for impedance control. 
Then VLM infers task-relevant stiffness and damping, grounding semantic understanding in physical interaction for safer, more adaptive manipulation.

\subsection{Leveraging VLMs for VIC controller}
Recent work has explored interpreting task context and generating adaptive impedance parameters by VLM, but often focuses on specific task domains or relies on simplified models that do not fully leverage the capabilities of modern VLMs\cite{batool2025impedancegpt, 11128409,11106445,jekel2025visio}.  

Existing works can be categorized into two main approaches. First, some studies tackle contact-rich tasks using impedance controllers \cite{11128409,11106445} but without VLM enhancement, where the vision-language component only assists in high-level task planning rather than real-time impedance adaptation. Second, works such as \cite{batool2025impedancegpt} have integrated VLMs to improve VIC parameter selection, but their evaluation is limited to simple scenarios but non-contact-rich manipulation tasks. Additionally, some recent efforts introduce extra modalities to enhance physical interaction performance, for instance, \cite{jekel2025visio} proposes a VLM-driven teleimpedance control framework that utilizes eye tracking, while \cite{bi2025vla} incorporates tactile information for improved contact sensing.

In contrast, our work uses in-context learning to adapt parameters based on task-specific context, integrates real-time force/torque feedback to ensure safe physical interactions.

\section{Methodology}
\begin{figure}
    \centering
    \includegraphics[width=0.9\linewidth]{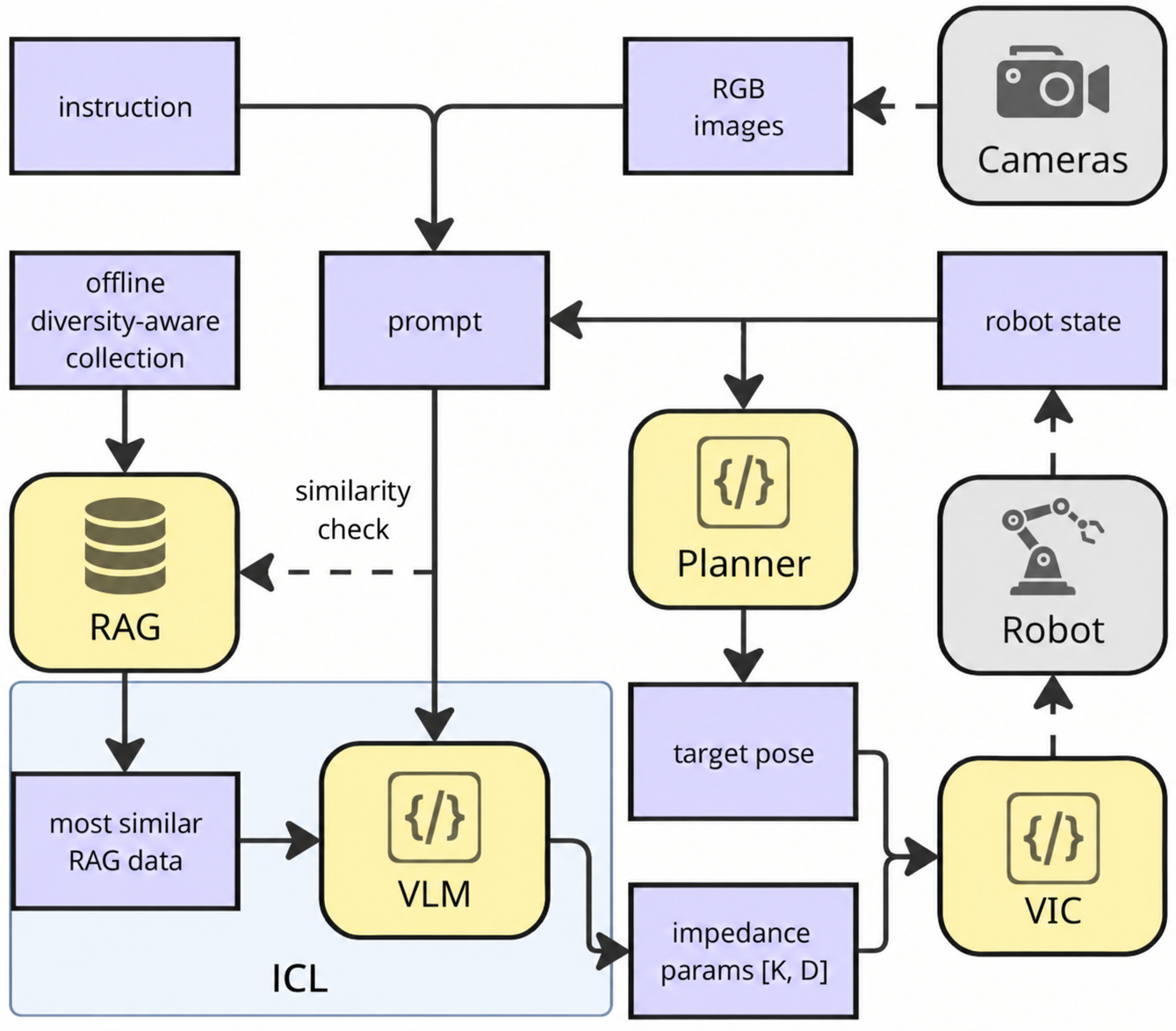}
    \caption{Overview of the proposed OmniVIC. 
    The system integrates a VLM with a VIC to enable safe and adaptive manipulation in contact-rich tasks.
    The VLM processes multimodal inputs, including visual observations, natural language instructions, and real-time force/torque feedback, to generate context-aware impedance parameters (stiffness and damping) for the VIC.
    }
    \label{fig:overview}
\end{figure}

In this section, we describe the overall architecture and key components of OmniVIC (Fig.~\ref{fig:overview}), 
a universal VIC empowered by RAG+ICL and assisted by VLM.
This integrates high-level semantic reasoning from multimodal inputs with low-level compliant control, 
enabling robots to adapt their impedance parameters for safe and effective manipulation in contact-rich scenarios. 

We detail the mechanism of VLM-informed parameter generation for the classical variable impedance control, the RAG process for leveraging prior experiences, the use of ICL for adaptive gain synthesis, 
and the real-time force/torque feedback regulation that ensures safe physical interaction. 
Implementation details are provided to facilitate reproducibility and highlight the practical considerations of deploying OmniVIC in both simulated and real-world environments.

\subsection{VLM-informed VIC}
A classical Cartesian VIC modulates the robot's stiffness $\boldsymbol{K}$ and damping $\boldsymbol{D}$ to achieve compliant interactions with the environment. We implement VIC in Cartesian space.
The control law is typically defined in task space as:
\begin{equation}
\boldsymbol{F}_{ext} =  \boldsymbol{D}\dot{\Tilde{\boldsymbol{x}}} + \boldsymbol{K}\Tilde{\boldsymbol{x}},
    \label{eq:cartesian_impedance}
\end{equation}
where $\Tilde{\boldsymbol{x}}$ is the Cartesian pose error between the desired and actual pose $\boldsymbol{x}_d,\boldsymbol{x}\in\mathbb{R}^{6}$. 
Accordingly, $\dot{\Tilde{\boldsymbol{x}}}$ is the velocity error between the desired and actual end-effector velocity, $\dot{\boldsymbol{x}}_d, \dot{\boldsymbol{x}}\in\mathbb{R}^{6}$, respectively. 
The desired stiffness matrix $\boldsymbol{K} \in \mathbb{R}^{6 \times 6}$ is a diagonal matrix with variable terms as $\boldsymbol{K}$=$diag\{K_x, K_y, K_z, \epsilon K_x, \epsilon K_y, \epsilon K_z\}$, 
where $\epsilon$ is the coefficient for rotation elements to reduce and simplify parameter tuning. 
The desired damping matrix $\boldsymbol{D} \in \mathbb{R}^{6 \times 6}$ is also diagonal matrix with variable terms as $\boldsymbol{D}$=$diag\{D_x, D_y, D_z, \zeta D_x, \zeta D_y, \zeta D_z\}$,
where $\zeta$ is the coefficient for rotation elements.

The key to effective VIC lies in the appropriate selection of the impedance parameters $\boldsymbol{K}$ and $\boldsymbol{D}$. 
Traditional approaches often rely on manual tuning or predefined schedules, which may not generalize well across different tasks or environments. 

In our method, the VLM processes visual and language inputs to derive the task context, which is then used to generate adaptive impedance parameters for the VIC controller. 
To simplify the problem, the VLM outputs diagonal translational stiffness and translational damping elements, reducing the parameter space to three stiffness values $(K_x, K_y, K_z)$ and three damping values $(D_x, D_y, D_z)$.

To enhance context-aware VLM parameter generation, we employ a RAG mechanism to fetch relevant prior experiences from a structured memory bank, see Sec.~\ref{sec:rag-data} and \ref{sec:rag-retrieval}.
These experiences inform the VLM about similar past tasks, improving its ability to generate appropriate impedance parameters for the current scenario.
Then, we use ICL (Sec. ~\ref{sec:icl}) to condition the VLM on these retrieved examples and the current task context, allowing for dynamic generation of impedance parameters without retraining to alleviate heavy computation.
Finally, we incorporate real-time force/torque feedback into the VIC controller to ensure that interaction forces remain within safe thresholds, further enhancing safety during manipulation.

\subsection{RAG data acquisition and storage} \label{sec:rag-data}

RAG data are acquired by deploying any robot policy and applying two success criteria: a maximum allowable contact force $F_{\max}$ and a time budget $T_{\max}$. A trial is labeled \emph{failure} if the measured force exceeds $F_{\max}$ at any time or if the task time exceeds $T_{\max}$; otherwise, it is labeled \emph{success}. 
To fully automate logging, we employ a lightweight VLM that processes the instruction $T$ together with the current twist and wrench and returns the controller gains $(K,D)$. 

For each \emph{successful} execution, we append the following to the RAG database: (i) the instruction $T$, (ii) the end–effector twist $v_t
\in\mathbb{R}^6$ expressed in the world frame,(iii) the phase label $p_t$, (iv) the gravity–compensated wrench $w_t=[\,F_t,\,\tau_t\,]\in\mathbb{R}^6$ expressed in the world frame, and (v) the VLM-predicted controller impedance parameters $(K,D)$. 
Items (i)–(iii) are detailed below.

\textbf{Instruction}
We store the instruction in two complementary forms: natural-language instruction text, denoted \(T_{\text{text}}\) (used later for in-context learning; see Section~\ref{sec:icl}), and a precomputed sentence embedding, denoted \(T_{\text{emb}}\) (used for retrieval; see Section~\ref{sec:rag-retrieval}). 
The embedding is produced once at collection time by a text encoder and persisted alongside the raw text so that retrieval does not require re-encoding. 
Throughout the paper, \(T_{\text{text}}\) refers to the raw instruction string and \(T_{\text{emb}}\) refers to its embedding.

\textbf{Twist \& wrench}
Unless otherwise stated, the 6D twist $v_t$ and the 6D wrench $w_t$ are expressed in the world frame via the standard $SE(3)$ adjoint frame transformation; the wrench is gravity-compensated prior to storage to isolate external/contact forces.

\textbf{Phase label}
By querying a VLM with the instruction $T$, the current image $I_t$ captured by an overview camera, and the world-frame twist $v_t$ and the world-frame wrench $w_t$. 
the model outputs one of $\{\text{Free\_motion},\text{Approaching},\text{Contact},\text{Retreat}\}$ as the phase label at each step.

When a new successful record arrives: (i) if the bank is not full, we add it; (ii) if the bank is full, we temporarily pool the new record with the records belonging to this instruction. 
Then, computing pairwise similarities using the method defined in Sec.~\ref{sec:rag-retrieval}, identify the closest pair (i.e., the two most similar records), 
and randomly discard one of the two. 
The remaining records are kept. This closest-pair replacement prevents near-duplicate motions from accumulating and preserves diversity under a fixed memory budget. The bank's update mechanism enables the RAG to improve its performance with continuous use.

\subsection{RAG retrieval}
\label{sec:rag-retrieval}
RAG retrieval process that identifies the most relevant prior experiences from the database to inform impedance parameter generation for the current request.
Given a new request, pre-processing is in a manner consistent with the storage pipeline: (i) the instruction raw string $T_{\text{text}}$ (for in-context learning Sec.~\ref{sec:icl}) and its embedding $x_{\text{emb}}$ computed with the same text encoder used in Sec.~\ref{sec:rag-data}; 
(ii) the phase label predicted by the VLM using $x_{\text{text}}$, the current image $I_t$, and the proprioceptive signals (twist $v_t$ and wrench $w_t$);
No re-encoding of database instructions is required because their embeddings were precomputed during collection. 

Aiming to reduce the search space and improve retrieval efficiency, the pre-processed items are then used in a four-step progressive retrieval process:

\subsubsection{Step 1: instruction filtering} \label{sec:step1}
We compare the query’s instruction embedding $T_{\text{emb}}$ to all stored instruction embeddings using cosine similarity, and retrieve the top-$M\%$ most similar instructions by this score.

\subsubsection{Step 2: phase filtering} \label{sec:step2}   
As each instruction in the RAG store is partitioned into four phase-specific records, 
we refine the results of Step 1 (Sec.~\ref{sec:step1}) by keeping only those records whose phase matches the query's phase for each of the top-$M\%$ instructions. 
The phase-consistent candidate pool is then used in Sec.~\ref{sec:step3} for similarity computation.

\subsubsection{Step 3: similarity computation}  \label{sec:step3}
We compute four modality-specific similarity scores between the query and each candidate record in the pool from Step 2 (Sec.~\ref{sec:step2}): (1) force similarity, (2) torque similarity, (3) linear velocity similarity, and (4) angular velocity similarity.
All four similarities use the cosine similarity, which returns a score in $[-1,1]$, as defined below:
\begin{equation} \label{eq:cosine-similarity}
\mathrm{cos\_sim}(a,b) = \frac{a \cdot b}{\|a\|\|b\|} 
\end{equation}
where \(a\) and \(b\) are the two vectors being compared, and \(n\) is their dimensionality.


\subsubsection{Step 4: final score} \label{sec:step4}
Although wrenches and twists represent different physical quantities, in Sec.~\ref{sec:step3} we compute four modality-specific \emph{similarity} scores: force, torque, linear velocity, and angular velocity, each using the same well-known cosine similarity. 
Because these are similarity scores, they naturally lie on the common symmetric range \([-1,1]\), making them directly comparable across modalities. 
We then sum the four similarity scores to obtain a single aggregate measure for each candidate, and keep the top five highest-scoring candidates as exemplars for in-context learning.

After completing Steps~1--4, all remaining examples are used as in-context exemplars. 
For each retained example, we extract the instruction text ($T_{\text{text}}$), the phase label, the controller gains $(K,D)$, 
and the associated motion signals (world-frame twist and gravity-compensated world-frame wrench), 
and the overall similarity score from Sec.~\ref{sec:step4}. These items are then formatted as in-context examples and concatenated with the current request to form the prompt for the VLM. 
By leveraging RAG, we ensure that the VLM is informed by relevant prior experiences, enhancing its ability to generate context-aware impedance parameters for the current task.

\subsection{ICL‑enhanced impedance parameter prediction}
\label{sec:icl}
Given a new request, we run RAG retrieval to obtain the top–$N$ most similar records. We then provide the top five similar records to the VLM as examples. Each example includes the instruction text ($T_{\text{text}}$), phase label, controller gains $(K,D)$, world-frame twist and wrench, and the similarity score. 
The VLM outputs $(K,D)$ for the current step using the prompt (simplified, see detailed in our code) below. Given bounded, filtered gains and a bounded update rate, the closed-loop remains stable under standard VIC assumptions.

\begin{tcolorbox}[fontupper=\ttfamily\small, 
    title=Simplified Prompt for OmniVIC impedance parameter dynamic generation:]
    
You are an expert robotic impedance controller capable of analyzing visual 
scenes and physical interaction states.

Given the \textcolor{red}{\{instruction\}},  
\textcolor{red}{\{current phase\}} , 
\textcolor{red}{\{twist\}}, and \textcolor{red}{\{wrench\}}, 
\textcolor{red}{\{impedance range\}} , 
determine optimal impedance parameters.

Apply phase-based impedance principles: 
\begin{itemize}
    \item Highest for free motion (precise control)
    \item Lowest for contact (maximum compliance)
\end{itemize}

Consider motion direction adaptation: 
\begin{itemize}
    \item Increase stiffness in the primary motion direction when overcoming resistance.
    \item Decrease stiffness in the primary motion direction when maintaining accuracy.
\end{itemize}

Reference similar successful example with \textcolor{red}{\{similarity score\}}:
\begin{itemize}
    \item Task: \textcolor{red}{\{example instruction\}}
    \item Phase: \textcolor{red}{\{example phase\}}
    \item Twist: \textcolor{red}{\{example twist\}}
    \item Wrench: \textcolor{red}{\{example wrench\}}
    \item Parameters: \\
          $K$ = \textcolor{red}{\{example K\}}, 
          $D$ = \textcolor{red}{\{example D\}}
\end{itemize}

\textbf{Output:} $K = [K_x, K_y, K_z]$, $D = [D_x, D_y, D_z]$

\end{tcolorbox}


\section{Experiments}
We evaluate OmniVIC on a suite of contact-rich manipulation tasks in both simulation and real-robot experiments. 
The evaluation metrics include task success rates, force violations, and overall interaction safety.
Through these experiments, we aim to answer the following research questions:
\begin{itemize}
    \item Does OmniVIC improve task success rates in physical manipulation tasks compared to baselines?
    \item How effectively does OmniVIC reduce force violations and enhance safety during manipulation?
    \item What is the impact of RAG and ICL on the adaptability and generalization of impedance parameters across diverse tasks?
\end{itemize}

\subsection{Simulation experiments}
\subsubsection{Baselines and Evaluation Tasks}

\textbf{Baselines}: 
As OmniVIC is a general-purpose variable impedance controller rather than a specific policy, we adapt robot policies from a pretrained VLA model Pi0~\cite{black2024pi_0} to execute tasks.
The default controller used in Pi0 is our baseline, which is a low-level position controller. Where the default parameters are fixed transitional stiffness $k_p = 150 N/m$, and damping $K_d = 2 \sqrt{k_p} =24.494 N s/m$. We compare the baseline with OmniVIC as our method.

\begin{figure}[!t]
    \centering
    \includegraphics[trim=0.1cm 2cm 0.1cm 0.5cm,width=0.97\linewidth]{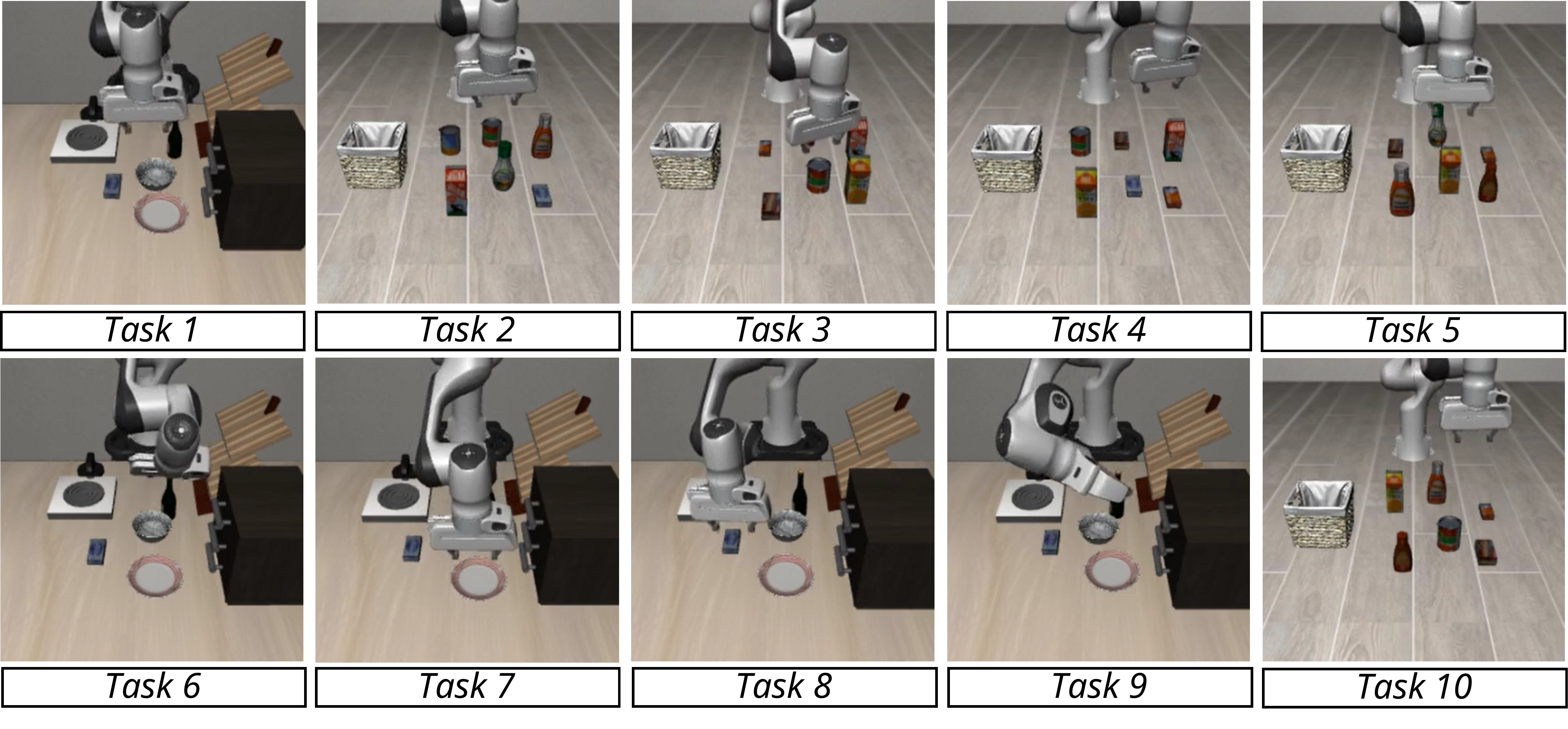}
    \caption{Tasks used in simulation experiments.}
    \label{fig:whole_task}
\end{figure}

\textbf{Evaluation task and protocol}:
We select 20 related manipulation tasks from LIBERO benchmark, 10 from \textit{LIBERO-Object} and 10 from \textit{LIBERO-Goal}.
By running the original benchmark under a force constraint, we design the split by stratified selection based on its outcome.
The split is composed of \textbf{Query Set} and \textbf{Knowledge-Base Set} to ensure evaluation tasks in \textbf{Query Set} are genuinely unseen in the retrieval component \textbf{Knowledge-Base Set}, avoiding train–test leakage and matching zero-shot evaluation practice in information retrieval.
To do so, we stratify the original outcome into bands so that tasks are represented evenly. 
From these bands, we form two disjoint sets of equal size: a \textbf{Knowledge-Base Set} of 10 tasks (retained solely for building the RAG store and never used as the later query) and a \textbf{Query Set} of 10 tasks (held out solely for issuing queries). 
The tasks for ICL evaluation are in \textbf{Query Set} with specific details illustrated in Fig.~\ref{fig:whole_task} and Table~\ref{tab:whole_task}, providing diverse manipulation scenarios with rich interactive behaviors. 

Safety/termination rule: As we care for safe contact, the contact force is monitored during execution. 
In addition to the task-specific success criteria defined by the policy, 
beyond the policy's native success criteria, an episode is marked as \emph{failure} if the measured interaction force exceeds \textbf{30\,N} on three consecutive steps to avoid sensor noise spikes.
We keep this safety rule active for all methods to ensure a fair comparison.



\begin{table}
\caption{Tasks in simulation experiments}
    \centering
    
    \begin{tabular}{cl}
    \hline
    \hline
        Task No. & Descriptions \\
        \hline
       Task 1 & put the bowl on top of the cabinet\\
       Task 2 & pick up the salad dressing and place it in the basket\\
       Task 3 & pick up the tomato sauce and place it in the basket\\
       Task 4 & pick up the milk and place it in the basket\\
       Task 5 & pick up the orange juice and place it in the basket\\
       Task 6 & put the wine bottle on top of the cabinet\\
       Task 7 & push the plate to the front of the stove\\
       Task 8 & put the cream cheese in the bowl\\
       Task 9 & open the middle drawer of the cabinet\\
       Task 10 & pick up the butter and place it in the basket\\
        \hline
    \end{tabular}
    
    \label{tab:whole_task}
\end{table}

\subsubsection{RAG data construction} \label{sec:rag_construct}
After stratified selection, we collect RAG data exclusively from the \textbf{Knowledge-Base Set}. To fully automate the pipeline, we pair the robot policy with a VLM-based assistance module that proposes task-dependent control parameters $(K, D)$ online. During $\pi_0$ rollouts on tasks in the Knowledge-Base Set, at each step the VLM receives the instruction text, the current phase label, and the previous-step world-frame twist and gravity-compensated world-frame wrench, and returns $(K, D)$. If the step completes without a force violation, we write a RAG record containing: the instruction as raw text $T_{\text{text}}$, its precomputed text-encoder embedding $T_{\text{emb}}$ (for retrieval), the phase label, the world-frame twist, the gravity-compensated world-frame wrench, and the controller gains $(K, D)$. Precomputing and storing both $T_{\text{text}}$ and $T_{\text{emb}}$ avoids re-encoding at query time and keeps indexing consistent with querying.

The instruction embeddings are computed with the BGE-M3 text encoder \cite{chen2024bge}. The VLM used throughout simulation is GPT-4o-mini.
The maximum capacity limit of our RAG database is designed as $B=20K$. In our experiments, we built the RAG dataset by running 10 tasks, with 20 records per task instruction–phase pair, resulting in a total capacity of 200 records for the evaluation.

\begin{figure*}[t!]
    \centering
    \begin{overpic}[trim=0.1cm 1cm 0.1cm 0.1cm,width=0.95\linewidth]{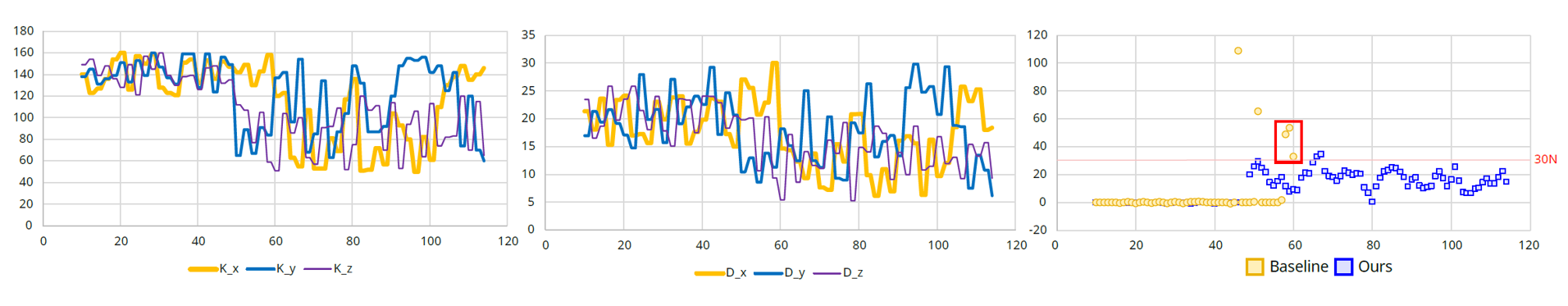}
    \put(2.5,14){\scriptsize{$K$ (N/m)}}
    \put(34.6,14){\scriptsize {$D$ } (N$\cdot$s/m)}
    \put(67.4,14){\scriptsize {$F_z$} (N)}
    \end{overpic}
    \caption{Deep dive into force and impedance profiles during task execution (Task 7 in Table~\ref{tab:whole_task}) in low-level controller. Left and middle: stiffness and damping profiles of OmniVIC. 
    Right: force (only $F_z$, the main contact force) profiles.
    The impedance profiles show that OmniVIC dynamically adjusts stiffness and damping based on task phases, reducing them during contact (middle area) to enhance compliance.
    The blue dots in force profiles demonstrate that OmniVIC maintains interaction forces within safe limits, avoiding spikes that
    violate constraints, while it happens (yellow dots in the red box) in the baseline controller and terminates the episode as a failure.
    }
    \label{fig:simulation_force_figure}
\end{figure*}

\begin{figure}[t!]
    \raggedright 
    \includegraphics[trim=0.1cm 2cm 0.1cm 0.1cm,width=0.45\textwidth]{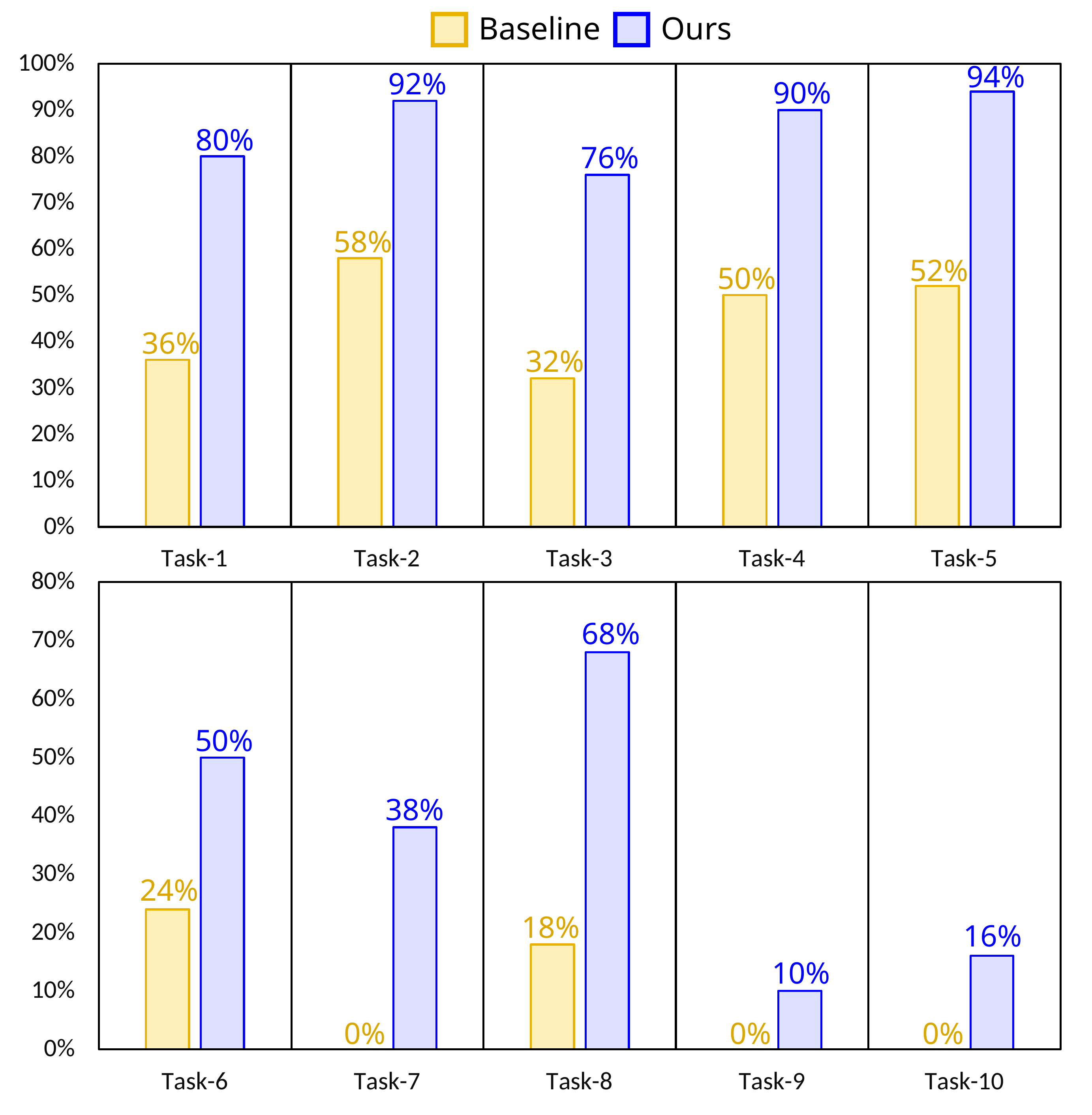}
    \caption{Simulation results: Task success rates of baseline ($\pi_0$ with position control) vs. OmniVIC (with VLM-informed variable impedance control).}
    \label{fig:accuracy_comparison}
\end{figure}

\subsubsection{Experiments evaluation for ICL} \label{icl-eval}
We conduct simulation experiments to validate the effectiveness of OmniVIC compared with the baseline in \textbf{Query Set}. 
All experiments were performed using 4 NVIDIA RTX A6000 GPUs.

We evaluate both the baseline and our OmniVIC-enhanced system on each of the 10 tasks in the \textbf{Query Set}, running 50 episodes per task per method. In all experiments, we set M in Step 1 (Sec.~\ref{sec:step1}) as $M=20\%$; consequently, Step 1 retains the top 20\% of instructions and Step 4 (Sec.~\ref{sec:step4}) retains the top five of the remaining candidates for ICL.

In the ICL evaluation, the original controller in Pi0 is the baseline, while our OmniVIC takes the top five similar examples retrieved from RAG bank with the current task context, and prompts them together to query VLM to generate the impedance parameters. 

\subsection{Real-world experiments}
We validate our approach on a physical robot platform to demonstrate its effectiveness in real-world contact-rich manipulation tasks.
\subsubsection{Experiment setup}
We implement OmniVIC on a Franka Emika Panda robot arm (at 1000 Hz), and set up a Logitech USB camera offering a global view of the robot arm and its workspace, and a 6-axis ATI Mini45 force/torque sensor for real-time interaction feedback. 
The VLM processes visual ($\sim$ 1 Hz) data from the camera and natural language instructions to generate adaptive impedance parameters (we set the stiffness range in $[300, 1000]$). 
The robot executes a series of contact-rich manipulation tasks, including drawer closing and object pushing, to evaluate the performance of our approach in real-world scenarios.


\begin{figure}[htbp]
    \centering
    \includegraphics[trim=0.17cm 3.2cm 0.1cm 0.1cm,width=0.45\textwidth]{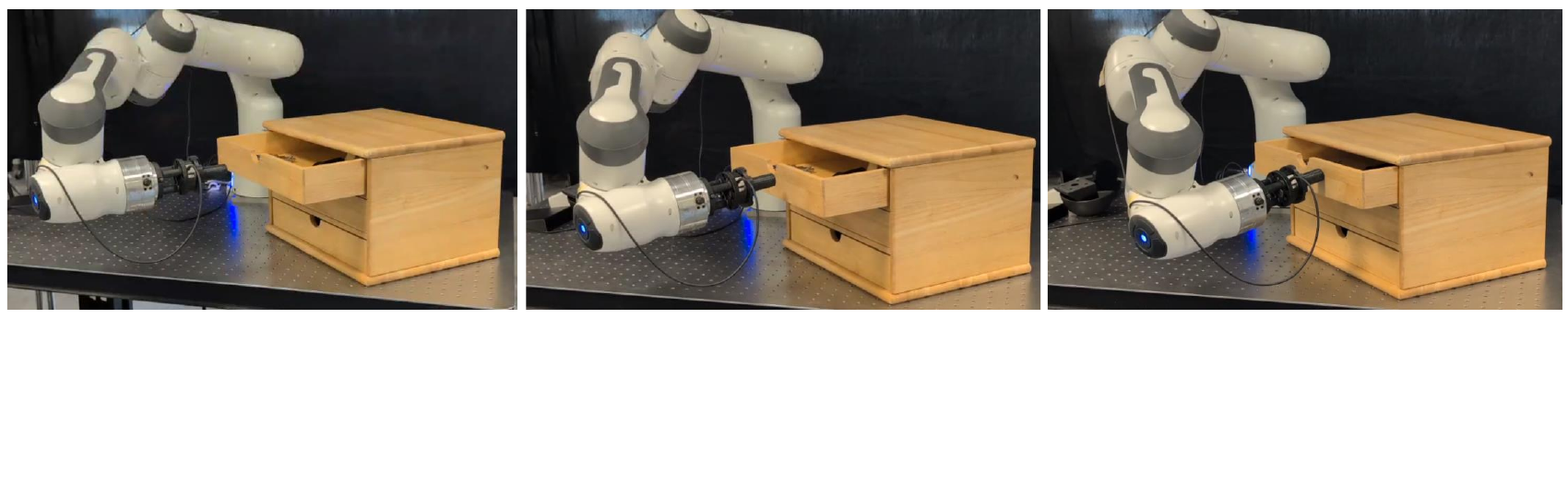}
    \\
    \includegraphics[trim=0.17cm 3.2cm 0.1cm 0.1cm,width=0.45\textwidth]{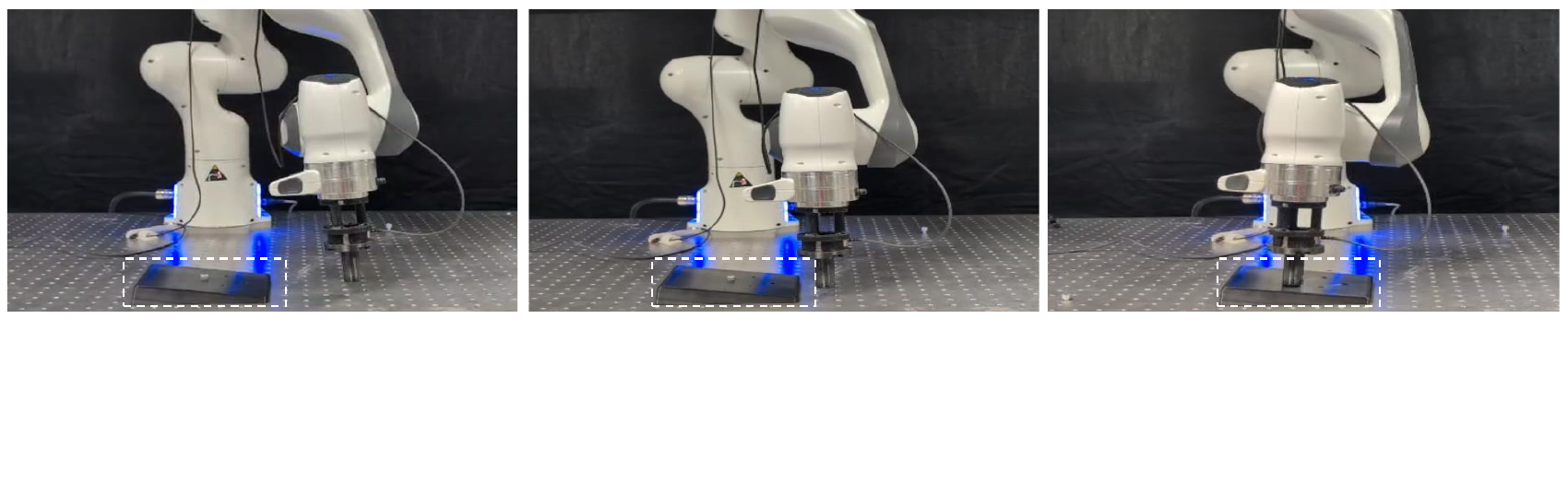}
    \\
    \includegraphics[trim=0.02cm 3.2cm 0.25cm 0.1cm,width=0.45\textwidth]{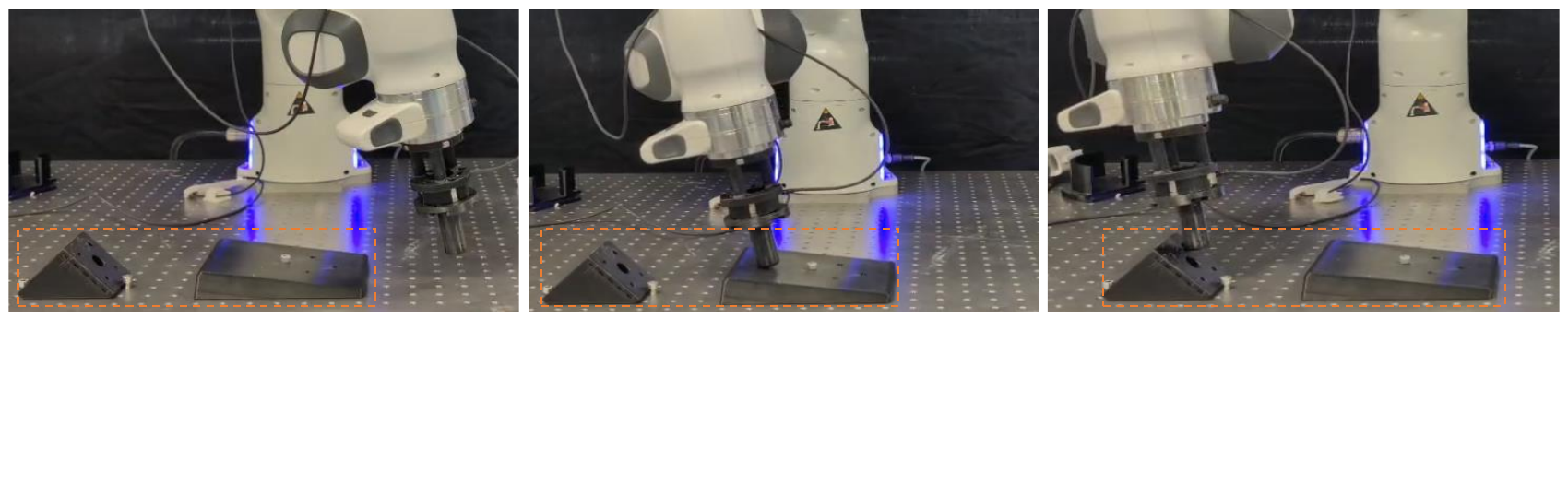}
    \caption{Real-world experiments: Top: Task 1: The robot gently closes the top drawer. 
    Middle: Task 2: The robot moves along the negative Y-axis while maintaining the same height in Z-axis, 
    where there is a ramp in the middle of the path to introduce contact.
    Bottom: Task 3: the same as Task 2, but with two different ramp shapes to test multiple phases of contact.
    }
    \label{fig:real_world_exp_setup}
\end{figure}



\subsubsection{Tasks descriptions}
We set up a series of contact-rich manipulation tasks on a physical robot. Each task is designed to test the robot's ability to adapt its impedance parameters based on visual and language inputs while maintaining safe interaction forces.
\begin{enumerate}
    \item Close the top drawer gently;
    \item Move along the negative Y-axis while keeping the same height in the Z-axis (one ramp);
    \item The same task setup as task-2, but with two ramps along the way.
\end{enumerate}
Moreover, in \textit{Task-1}, we design two comparison experiments: constant high $K \& D$ and constant low $K \& D$ to further validate the performance of OmniVIC. In these experiments, high and low stiffness values are set to 1000 and 300 N/m, respectively, while the damping is defined as $2 \times 0.707 \times \sqrt{k}$ accordingly, where $k$ indicates stiffness.

\section{Experiment Results}

\subsection{Simulation results and analysis}
We evaluate the task success rates of OmniVIC compared to the baseline. The results show that our approach achieves significantly higher success rates across all contact-rich manipulation tasks,
demonstrating the effectiveness of integrating RAG-ICL enhanced VLMs with VIC for safe and compliant interactions. 

\subsubsection{Force safety analysis}
Fig.~\ref{fig:simulation_force_figure} illustrates the critical advantage of our approach in maintaining safe interaction forces from one episode of the task \texttt{push the plate to the front of the stove}. Our OmniVIC dynamically maintain contact forces within a safe threshold of 30N. In contrast, the baseline without OmniVIC repeatedly violates force constraints, demonstrating unsafe interaction behaviors that could lead to task failure or equipment damage. 

As detailed in Fig.~\ref{fig:simulation_force_figure} (right), during the initial execution phase, both baseline and OmniVIC maintain low interaction forces as the robot approaches the target plate to push.
Then, OmniVIC reduces stiffness to allow gentle engagement with the target, minimizing impact forces. As the task progresses, it increases stiffness to ensure precise control while still adhering to force limits. 
This dynamic adjustment is crucial for tasks requiring delicate handling, such as placing objects in confined spaces or interacting with fragile items. 
The baseline approach, lacking this adaptability, often applies excessive force during these phases, leading to violations of safety constraints. It failed at step 60 in this example.

\subsubsection{Task performance evaluation}
We evaluate task performance across all 10 simulation scenarios, measuring both task success rates and adherence to the force constraint. Task success is defined strictly in terms of safety: any violation of the force constraint results in an immediate failure.
As shown in Figure~\ref{fig:accuracy_comparison}, the integration of OmniVIC yields substantial improvements across all evaluated tasks. 
The baseline achieves an average success rate of 27.0\% across tasks, with frequent force violations compromising both safety and task completion. 
In contrast, the OmniVIC-enhanced system achieves an average success rate of 61.4\%, a 2.27-fold improvement over the baseline. 
The success of OmniVIC stems from its ability to intelligently adapt impedance parameters based on task context. 
Through the RAG component, the system retrieves relevant prior experiences from similar manipulation scenarios, 
while the ICL mechanism enables real-time parameter generation tailored to the current task state. 
This dual mechanism allows for nuanced control strategies, applying higher stiffness during precise positioning phases while reducing stiffness during contact establishment to prevent excessive forces.


Moreover, the adaptive impedance control allows for smoother transitions between different motion phases (Fig.~\ref{fig:simulation_force_figure}), minimizing abrupt force changes that could lead to task failures or damage to objects. 
Consecutively, OmniVIC still maintains success rates even when the baseline fails due to force violations (Task7 - Task10 in Fig.\ref{fig:accuracy_comparison}), highlighting its robustness in handling complex contact-rich tasks. 

\subsubsection{Ablation study: RAG vs.~RAG+ICL (OmniVIC)} To quantify the contribution of ICL on top of RAG, we compare two configurations under identical settings: \textbf{RAG-only}, which applies the retrieved parameters without ICL, and \textbf{RAG+ICL (OmniVIC)}, which feeds the retrieved exemplars to the VLM for in-context learning to predict the impedance parameters. We evaluate both cases on \emph{Tasks~1--10} in simulation. Across the ten tasks, \textbf{RAG+ICL} achieves a \emph{1.59-fold} higher task success rate than \textbf{RAG-only}, indicating a substantial benefit from letting the VLM adapt the retrieved priors to the instantaneous scene and interaction state. Qualitatively, RAG-only often fails in contact-rich phases where the retrieved examples are similar but not fully representative of the current contact geometry or dynamics; the VLM with ICL compensates by \emph{contextualizing} those priors, yielding better outputs.
This ablation supports our design choice to pair RAG with ICL for robust, fine-grained modulation of $(K,D)$ in unseen and complex scenarios.

\subsection{Real-world results and analysis} \label{sec:real-exp}
We validate the effectiveness of OmniVIC in three physical interaction 
tasks as shown in Fig.~\ref{fig:real_world_exp_setup}. 
Each experiment demonstrates the robot's ability to adapt its impedance parameters based on visual and language context while maintaining safe interaction forces.
We measure task success rates, force violations, and overall interaction safety during task execution.

\begin{figure}
    \centering
    \begin{overpic}[
        trim=0.1cm 1.2cm 0.1cm 0.1cm,
        clip,
        width=0.9\linewidth
    ]{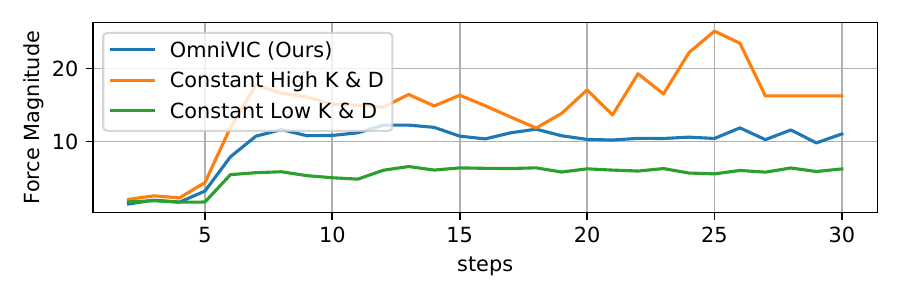}
        \put(5,22){\makebox(0,0)[l]{\scriptsize (N)}}
    \end{overpic}
    \\
    \begin{overpic}[
        trim=0.1cm 1.2cm 0.1cm 0.1cm,
        clip,
        width=0.9\linewidth
    ]{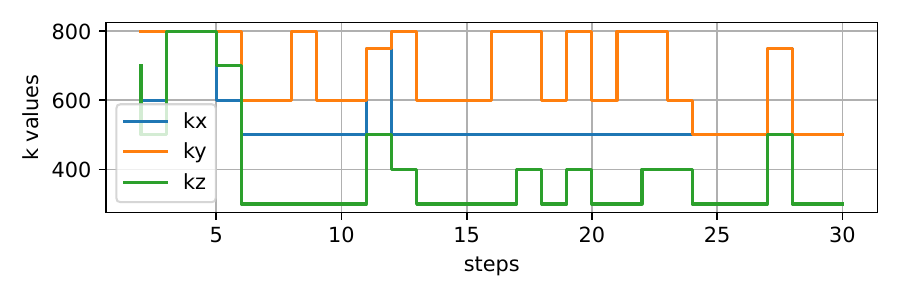}
    \put(5,25){\makebox(0,0)[l]{\scriptsize (N/m)}}
    \end{overpic}
    \\
    \begin{overpic}[
        trim=0.1cm 0.8cm 0.1cm 0.1cm,
        clip,
        width=0.9\linewidth
    ]{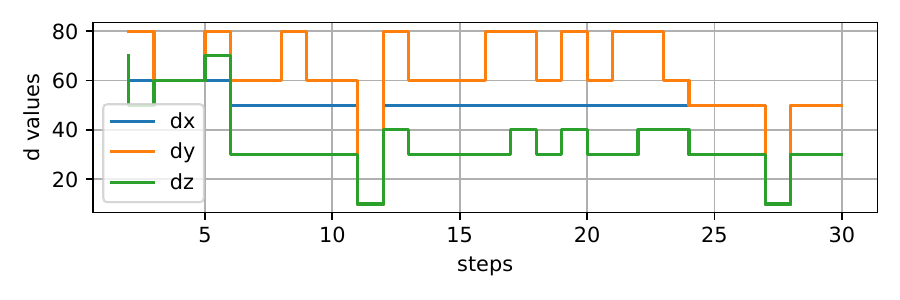}
    \put(5,28){\makebox(0,0)[l]{\scriptsize (N$\cdot$s/m)}}
    \put(98,2){\makebox(0,0)[l]{\scriptsize (step)}}
    \end{overpic}

    \caption{Task-1: The robot gently closes the top drawer using OmniVIC variable impedance control. The system adapts stiffness and damping parameters based on visual and language context to ensure safe interaction and avoid excessive force.}
    \label{fig:exp1_force}
\end{figure}

Starting with Task-1 (Fig.~\ref{fig:exp1_force}), we illustrate how OmniVIC effectively regulates interaction forces while closing the drawer gently without safety constraint violation. 
The force stays within safe limits throughout task execution, demonstrating the approach’s real-world suitability. In contrast, the comparison experiments in Fig.~\ref{fig:exp1_force} (top) show that constant high impedance causes excessive force, while constant low impedance fails due to insufficient force to overcome friction. Our method achieves a 100\% success rate across five trials, with zero force violations in all task scenarios.

\begin{figure}
    \centering
    \begin{overpic}[
        trim=0.45cm 0.7cm 0.1cm 0.1cm,
        clip,
        width=0.91\linewidth
    ]{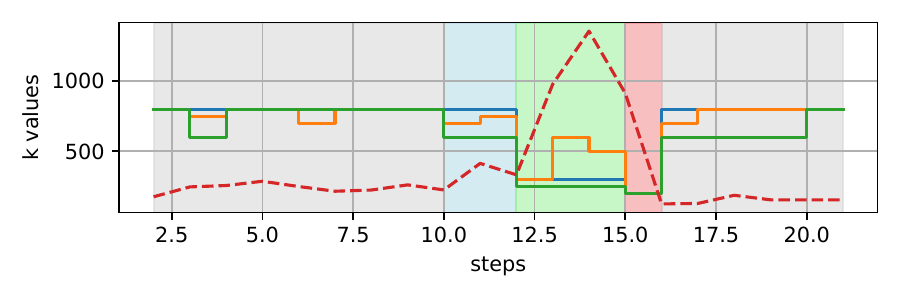}
    \put(3,25){\makebox(0,0)[l]{\scriptsize (N/m)}}
    \put(93,3){\makebox(0,0)[l]{\scriptsize (step)}}
    \end{overpic}
    \\
    \begin{overpic}[
        trim=0.1cm 0.18cm 0.1cm 0.2cm,
        clip,
        width=0.95\linewidth
    ]{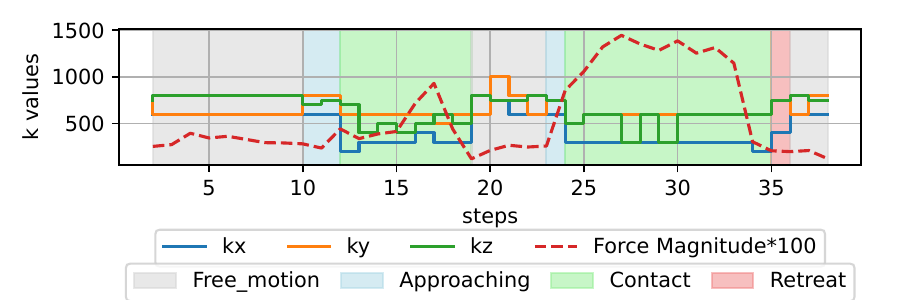}
    \put(5,32){\makebox(0,0)[l]{\scriptsize (N/m)}}
    \put(90,10){\makebox(0,0)[l]{\scriptsize (step)}}
    \end{overpic}
    \caption{Task‑2 (top): The robot moves along the negative Y‑axis while keeping a fixed Z‑height, with OmniVIC adjusting stiffness and damping from visual–language cues to maintain safe, low‑force interaction.
Task‑3 (bottom): The robot executes a sequence of tasks, continuously adapting impedance parameters from visual–language context to avoid excessive force.
Shaded regions denote contact phases detected by OmniVIC.
    }
    \label{fig:real_exp2}
\end{figure}

Furthermore, we elaborate how the stiffness has been regulated according to the task execution \textit{phase} in 
Task-2 and Task-3 As shown in Fig.~\ref{fig:real_exp2}, we further validate the adaptability of OmniVIC in more complex scenarios involving continuous contact and multiple phases of interaction. 
The robot successfully navigates along a path with a ramp, maintaining safe interaction forces while adapting its impedance parameters to the varying contact conditions. 
The stiffness and damping parameters are adjusted appropriately in real-time based on the visual and language context, ensuring smooth transitions between different phases of contact. 
The force remains within safe limits throughout the task execution, demonstrating the robustness and effectiveness of our approach in handling complex real-world manipulation tasks.

\section{Conclusion}
We present OmniVIC, a variable impedance controller that integrates RAG and ICL with VLMs to adapt impedance parameters to task and environmental contexts. Extensive simulation and real-world experiments demonstrate improved success rates and fewer force violations, increasing average success from 27\% (baseline) to 61.4\%.
The integration of RAG allows the system to retrieve relevant prior experiences, while ICL enables the VLM to generate context-aware impedance parameters in real-time. 
This dual mechanism enhances the robot's ability to perform safe and compliant interactions in diverse manipulation scenarios.
Future work will investigate the generalization of our approach to a wider range of physical tasks extension to mobile robots and humanoids.   

\bibliographystyle{IEEEtran}
\bibliography{main}
\end{document}